\documentclass[10pt, conference]{IEEEtran}
\usepackage[utf8]{inputenc}
\usepackage[left=1.0in,right=1.0in]{geometry}
\usepackage{setspace}
\usepackage{multirow}
\usepackage{graphicx}
\usepackage{tikz}
\usepackage{hyperref}
\usepackage{listings}
\usepackage{sidecap}
\usepackage{wrapfig}
\usepackage[labelfont=bf]{caption}
\usepackage{subcaption}

\hypersetup{
    colorlinks,
    citecolor=black,
    filecolor=black,
    linkcolor=black,
    urlcolor=black
}

\begin{document}

\title{An FPGA-Accelerated Design for Deep Learning Pedestrian Detection in Self-Driving Vehicles}

\author{\authorblockN{Abdallah Moussawi, Kamal Haddad, and Anthony Chahine}
\authorblockA{Department of Electrical and Computer Engineering\\
American University of Beirut\\
Beirut, Lebanon\\ 
Email: aam68@aub.edu.lb, kmh16@aub.edu.lb, atc07@aub.edu.lb}
}

\maketitle

\begin{abstract}
With the rise of self-driving vehicles comes the risk of accidents and the need for higher safety, and protection for pedestrian detection in the following scenarios: imminent crashes, thus the car should crash into an object and avoid the pedestrian, and in the case of road intersections, where it is important for the car to stop when pedestrians are crossing. Currently, a special topology of deep neural networks called Fused Deep Neural Network (F-DNN) is considered to be the state of the art in pedestrian detection, as it has the lowest miss rate, yet it is very slow. Therefore, acceleration is needed to speed up the performance. This project proposes two contributions to address this problem, by using a deep neural network used for object detection, called Single Shot Multi-Box Detector (SSD). The first contribution is training and tuning the hyperparameters of  SSD to improve pedestrian detection. The second contribution is a new FPGA design for  accelerating the model on the Altera Arria 10 platform. The final system will be used in self-driving vehicles in real-time. Preliminary results of the improved SSD shows 3\% higher miss-rate than F-DNN on Caltech pedestrian detection benchmark, but 4x performance improvement. The acceleration design is expected to achieve an additional performance improvement significantly outweighing the minimal difference in accuracy.\\
\end{abstract}

\IEEEpeerreviewmaketitle

\section{Introduction}

Self-driving vehicles are at the forefront of future transportation technology, nevertheless, the question of pedestrian safety always arises when discussing them. It is a question of whether or not autonomous vehicles can detect pedestrians, distinguish them from other objects and perform the right course of actions(s) accordingly.  
In this paper, the goal is to design, implement, and accelerate  a system that detects pedestrians and distinguishes between humans and objects, on a field programmable gate array (FPGA), with self-driving vehicles in mind.

\subsection{Motivation}

The motivation behind the project is to apply engineering disciplines in solving the problems of detecting humans vs. objects in self-driving vehicles.\\

According to the National Highway Traffic Safety Administration, approximately 94\% of vehicle-related accidents in the United States are caused by human drivers \cite{accidents}. Technology companies have removed the cause of this large percentage by building self-driving vehicles.\\

Research pertaining to pedestrian detection is vital to make driver-less cars safer, and more acceptable in society. This project will aid in self-driving cars becoming commodities as cars are nowadays, because they will inevitably save more lives. It is important for self-driving cars to avoid all obstacles, but with a higher priority on humans. For example, in a scenario where a vehicle had to crash into either an object or a human, it would crash into the object.





\section{Literature Review}
\subsection{Pedestrian Detection}
\subsubsection{Pipeline}

\begin{figure}[h]
\centering
\includegraphics[scale=0.3]{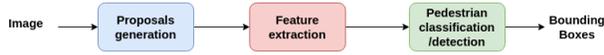}
\caption{Pedestrian detector pipeline}
\label{fig:pipeline}
\end{figure}

A common pipeline to build a pedestrian detector is as shown in Figure \ref{fig:pipeline}. First pedestrian bounding boxes are proposed, then image features are extracted from each proposed bounding box, and finally those features are passed to a classifier to classify each bounding box as a pedestrian or not.\\

Proposals generation can be done using a simple sliding window as used in \cite{hog}. In \cite{selecS}, they proposed a method called Selective Search that generates bounding boxes that most likely contain an object. \cite{frcnn} proposes a convolutional neural network called Region Proposal Network RPN that predicts the bounds of an object and its "objectness score" in an image.\\

People were mostly working on the feature extraction, by trying to find the features that most represents a pedestrian. Those features are either hand-crafted or generated by convolutional neural networks. More on that in Section \ref{sec:history} and \ref{sec:stateofart}.\\

For classification, in \cite{hog} a linear SVM is used, \cite{vj} uses a cascade of classifiers, and deep learning based methods uses fully connected layers as in \cite{ang}.\\

This is a general pipeline, some methods might combine proposals generation and feature extraction in a single box, others might combine feature extraction and classification, and it's also possible to do all of them in a single shot as in \cite{ssd}.\\

\subsubsection{History} \label{sec:history}
Pedestrian detection has been an on-going research for more than a decade. Its importance for surveillance systems, autonomous driving cars, robotics, has been driving many researchers to work on that problem. One of the first successful object detection methods was Viola and Jones \cite{vj} in 2001, it was first applied successfully on face detection, it was then applied to pedestrian detection, but it had a high miss-rate (95\%) on the challenging Caltech dataset \cite{caltech}. Later on in 2005, a more successful method, at its time, called Histogram of Oriented Gradients HOG, was published \cite{hog}, it uses a histogram of oriented gradients as a feature set and it achieves a low miss-rate (46\%) on the easy pedestrian INRIA dataset, but it has a high miss-rate (68\%) on Caltech dataset \cite{caltech}. In \cite{chftrs}, they use a bigger set of features called Integral Channel Features (ChnFtrs) that are a combination of multiple channels features such as color (RGB, grayscale..etc), gradient histogram(i.e HOG), image gradients magnitude. It had a miss-rate (56\%) on Caltech dataset \cite{caltech}. All theses methods followed the same pipeline. For proposing regions, all used a simple sliding window approach, then those windows are processed for feature extraction (HOG, Haar, wavelets..etc), and for classification, some used cascaded decision trees classifiers with Adaboost training(Viola Jones, ChnFtrs) and other used support vector machines SVM (HOG). Other methods also existed, many of them were variants of the above methods, where the researchers worked on better set of features and better classifiers. 


\subsubsection{State Of The Art} \label{sec:stateofart}
Due to advancements in deep learning, specifically the convolutional neural networks used for image classification and object detection \cite{alexnet} \cite{frcnn}, many of the state-of-the-art pedestrian detection methods are using ConvNets as a box in the pipeline to either do regions proposals, feature extractions, classification, or any combination of them. For instance, Angelova et. al \cite{ang} proposed a deep network called Large-Field-Of-View (LFOV) to generate proposals that might contain pedestrians and then pass those proposals to AlexNet \cite{alexnet} for pedestrian verification and they achieved a relatively low miss-rate (35\%) and a speed of 4FPS. Also Angelova et al. in \cite{cascade} used a cascade of the previously mentioned deep network \cite{ang} and a cascade of VeryFast \cite{vf}, and they were able to achieve with some tuning a miss-rate of 26.21\% with an almost real-time speed of 15FPS. In \cite{rpnbf}, researchers use the Regional Proposal Network from \cite{frcnn} for proposal generation, which does very well as a standalone pedestrian detector with 14.9\% miss-rate, and use boosted forests for better classification and verification of the generated proposals to achieve a competitive miss-rate of 9.6\% but at a low frame-rate of 2FPS. In \cite{fused}, authors proposed a method called Fused-DNN where they use the Single-Shot MultiBox Detector \cite{ssd} as a proposal generation box, and use a finetuned GoogleNet and ResNet models for pedestrian verification, by retraining these models on pedestrian dataset, and they were able to achieve the lowest miss-rate on Caltech dataset of 8.65\% at 6FPS. All evaluations of the above literature were done on the Caltech test dataset by their corresponding authors and tested on high-end GPUs.\\


\subsection{FPGA Acceleration Of Convolutional Neural Networks}

\begin{figure}[h]
\centering
\includegraphics[width=0.5\textwidth]{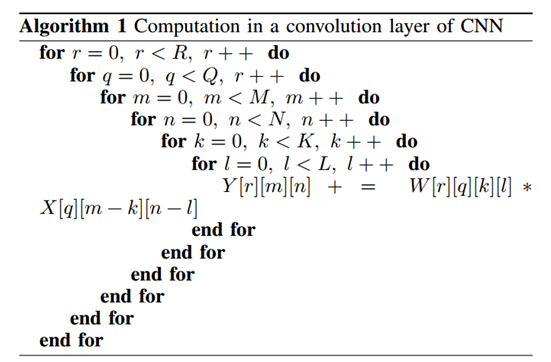}
\caption{Computation in a convolutional layer of a CNN \cite{accelhz}}
\label{fig:convcode}
\end{figure}

As discussed above, neural networks are highly accurate in detecting people, but they are also computationally intensive. 90\% of the computations of AlexNet, for example,are in the convolutional layers \cite{accelhz}. Figure \ref{fig:convcode} shows the computation of a single convolutional layer of a convolutional neural network.\\

ZynqNet \cite{zynqnet} accelerates not just the convolutional layers of SqueezeNet \cite{squeezenet} but also the ReLU nonlinearities, concatenation, and the global average pooling layers on the Zynqbox, which includes a Xilinx Zynq XC-7Z045 SoC, 1 GB DDR3 memory for the ARM processor, 768MB independent DDR3 memory for the programmable logic (PL), and a 1 GHz CPU is connected to the PL via AXI4 ports for data transfer.\\


ZynqNet \cite{zynqnet} runs at a frame rate of 0.51FPS, with image sizes of 227x227 with the processor running at 100MHz instead of the maximum which is 200MHz for some undisclosed reason. The claim is that with performance enhancements such as implementing 1x1 convolutions more efficiently, removing the global pooling cache, fixing the architectural bottleneck seen in pre-fetching images from the image cache, and switching to 16 bit fixed point data format \cite{zynqnet} could potentially boost the frame rate to 30FPS, and even more improvements could be made on the network to reach 62FPS. The usage of floating point came as a result of using Caffe framework \cite{caffe}, whose weights are represented in floating point format. \\

Concerning power, ZynqNet \cite{zynqnet} runs at 0.53 images/Joule and dissipates 7.8W with the accelerator running, and 7.46W while the accelerator is idle. But with the improvements, the efficiency increases to approximately 5.4 images/Joule, which is higher than the efficiencies of both the Intel Core i7 CPU (1.3 images/Joule), and the Nvidia Titan X (2.5 images/Joule), but less than the efficiency of the Nvidia Tegra X1 (8.6 images/Joule) \cite{zynqnet}.\\






In \cite{optimizing}, communication and computation were identified as the 2 main constraints on throughput. The attainable performance is defined as the minimum of the computational roof, which is the peak throughput, and memory bandwidth * the computation to communication ratio, which is the DRAM traffic needed by a single kernel.\\

\begin{figure}[h]
\centering
\includegraphics[width=0.3\textwidth]{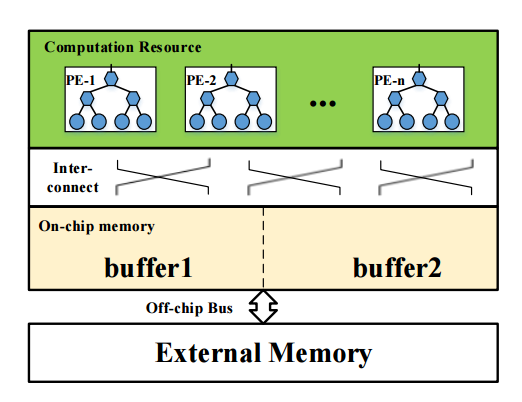}
\caption{FPGA Accelerator Design of \cite{optimizing}}
\label{fig:desgin}
\end{figure}


Loop tiling is a parallelization technique used to partition a loop into different loops and executed the loop in chunks, the reason this is usually is because it is sometimes unable to load the entirety of the data directly into the buffer. This method is used to fit data onto the limited caches of the Xilinx VC707 FPGA, where data that is too large for the cache is retrieved in blocks so a certain element in a loop is now reused after $\frac{iterations}{block size}$ iterations instead of n iterations. Loop unrolling, which was discussed before and loop pipelining are also methods used to increase parallelism. Loop pipelining enables the use of overlapped instructions by running the iterations in a pipelined fashion. Only the convolutional layers of AlexNet \cite{alexnet} were executed on the FPGA in \cite{optimizing}. This FPGA accelerator achieved a 4.74x speed-up with respect to a 2.2GHz CPU running with 16 threads in parallel.

\section{Design}
Field Programmable Gate Arrays (FPGAs) are used for accelerating CNNs, for their scalability and power efficiency compared to GPUs that are more powerful but also more expensive. In this project power and cost were important factors when designing the system for acceleration.\\

The design for accelerating the neural network on the high-end Altera Arria 10 is based on a previous VHDL design of AlexNet on the Xilinx ZedBoard. We are currently writing the VHDL code layer by layer, starting from the first layer.\\

Since the FPGAs have a limited number of DSP blocks and a limited amount of Random Access Memory (RAM), the layers must be tiled, meaning that layers must be loaded to the FPGA from external memory in chunks, similar to what is shown in Figure \ref{fig:desgin}. As the network propagates forward, new layers are loaded, but it important that if layers that are already on the board need to be reused, then no fetching from external memory will be done. Improvements in convolutional computations \cite{optimizing}, and the pipelining and parallelism found in \cite{zynqnet}, and \cite{optimizing} will also help accelerate the network to run in real-time.\\

\begin{figure}[h]
\centering
\includegraphics[scale=0.3]{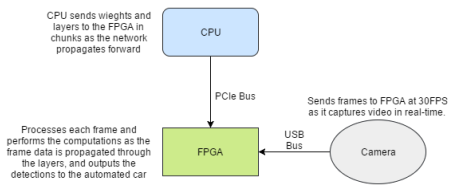}
\caption{System Block Diagram}
\label{fig:systembd}
\end{figure}

Figure \ref{fig:systembd} shows a block diagram of the system from a high-level perspective, and summarizes the procedure of detecting a pedestrian in real-time.

In our design, we are limiting the number of memory accesses by exploiting the DDR3 DRAM spatially by placing all coefficients next to one another. Additionally, we are reducing the number of MACs (Multiplication and Addition Computations) by storing some computation results in buffers and reusing them for other computations. The design also does not use a hard processor, but instead uses the Nios II softprocessor provided by Altera which we program to initialize DMA transactions, and place the coefficient values next to each other in memory, and as the frames are inputted from the camera into the FPGA, they are also stored in memory, and requested when computations are needed to be done on them. Therefore, no communication between the processor and the DRAM is needed, the memory reads and writes are directly done through a soft memory controller, though we will transition to a hard memory controller because it provides almost double the bandwidth.\\

We are also quantizing our caffe model, and testing, layer by layer, to find the optimal number of bits for fixed point computations such that the model's miss rate is slightly affected. This step is important because the Arria 10's DSP slices require fixed point multiplications and additions, and that significantly speeds up the model when compared to the 32 bit floating point computations that our GTX TITAN X is doing.

\section{Preliminary Implementation And Testing}
\subsection{Single-Shot MultiBox Detector SSD}

Single-Shot MultiBox Detector \cite{ssd} is a deep network based general object detector. It achieved a relatively high average precision of 74.3\% on PASCAL VOC general object detection competition at high speed of 59FPS \cite{ssd}.\\

In \cite{ssd}, the authors propose SSD, a fully convolutional neural network. SSD discretizes the possible output bounding boxes into a default set of bounding boxes at different scales and aspect ratios. The model predicts the object scores of each default bounding box, and regress the output bounding boxes' offsets to those default bounding boxes.\\

\begin{figure}[h]
\centering
\includegraphics[scale=0.22]{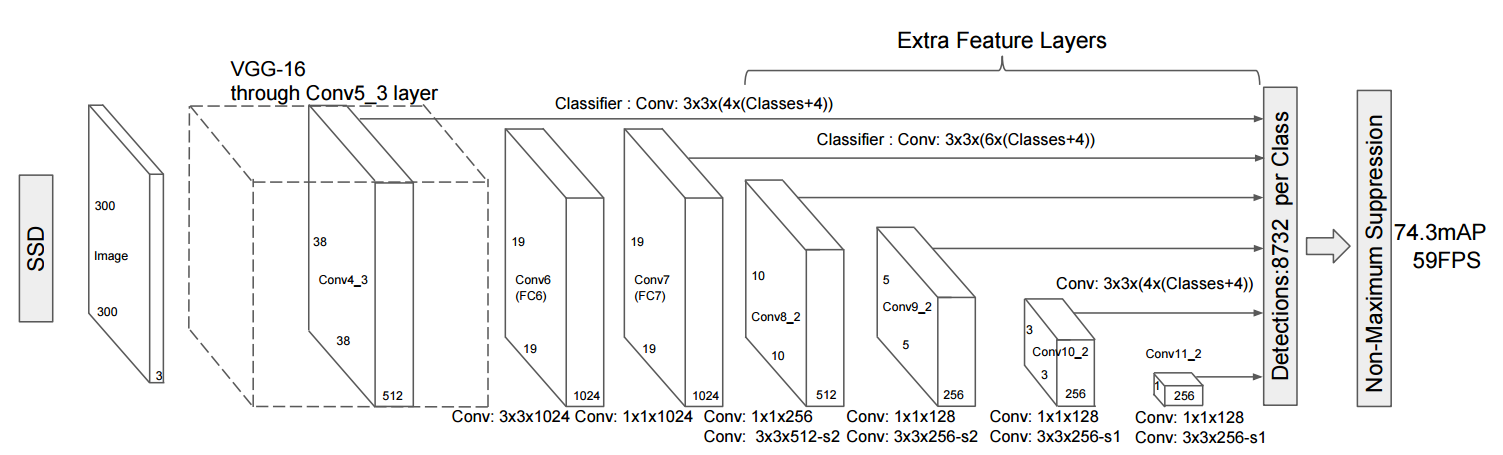}
\caption{SSD model \cite{ssd}.}
\label{fig:ssdm}
\end{figure}

\begin{figure}[h]
\centering
\includegraphics[scale=0.3]{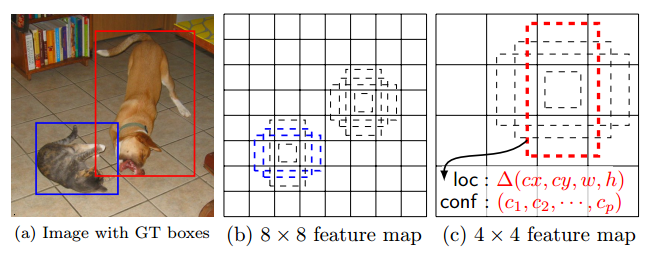}
\caption{Default boxes and groundtruth matching \cite{ssd}}
\label{fig:ssddb}
\end{figure}

SSD model architecture is shown in Figure \ref{fig:ssdm}. It uses VGG16 \cite{vgg} as a base network to generate feature maps. VGG-16 is is an ImageNet classification competition second place runner, and ImageNet localization competition first place runner. SSD also uses additional feature layers at the end of the base network as shown above. Some of the generated feature maps are passed to a convolutional predictors to compute the confidence of each default bounding box in each of these feature maps and regress the bounding box offsets. During training, the default bounding boxes are matched to the groundtruth bounding boxes as shown in Figure \ref{fig:ssddb}. The number of default boxes, their sizes and aspect ratios represent the hyper-parameters of the SSD model that should be chosen wisely based on the dataset objects sizes and aspect ratios.\\

\subsection{Training and Testing}

We have chosen Single-Shot MultiBox Detector \cite{ssd} as the baseline detector, as it was shown to achieve low miss-rate of 13.06\% on Caltech pedestrian dataset \cite{fused}. SSD is implemented using Caffe \cite{caffe} to train and test.\\

In order to train SSD model, we had to collect pedestrian dataset. Caltech pedestrian dataset \cite{caltech} is used for both training and testing. This dataset contains 10 hours of 30FPS video. This dataset is split into 6 sets for training and 5 sets for testing. It contains three annotations, 'person' if a pedestrian was identified, 'people' if a large group of pedestrians was identified, and 'person?' if it was unclear whether the identified object is a person or not. We only used 'person' annotation and any other annotation was ignored. For each video in the Caltech training dataset, we extracted one frame every 5 frames, that resulted in ~26k images, with a total of ~25k annotations of 'person'. In addition to Caltech dataset, we used both ETH pedestrian dataset \cite{eth} and TUD-Brussels datasets \cite{tudb}, that resulted in additional ~4k images, and ~14k annotations. The dataset preprocessing was done using the Caltech MATLAB evaluation/labeling code \cite{caltech}. We then setup the model with the hyper-parameters based on the statistics of pedestrians scales and aspect ratios of Caltech pedestrian dataset \cite{caltech}. After a lot of testing and experiments, we concluded that using a single aspect ratio [0.41] was enough to describe the default bounding boxes at all layers, and we used 3 different scales [4\%, 7\%, 8.5\%] at the con4\_3 layer, these scales corresponds to small size pedestrians, while the scales on the remaining layers were equally spaced starting with 0.1 on fc7 and ending with 0.9 on conv10\_2 layer. We also finetuned from a pre-trained SSD model that was trained on VOC+COCO general object detection datasets. The current configuration led to a state-of-art miss-rate of 11.88\% and running at 24FPS on NVIDIA GTX Titan X, which is best to our knowledge the fastest state-of-art model. Table \ref{tab:progress} shows the improvements we have done on the default configuration of SSD model. Table \ref{tab:comparison} compare our current model SSD with other state-of-art models. Overall performance stands for pedestrians that are at least 20pix tall with less than 80\% of their body occluded, while Reasonable performance stands for pedestrians that are at least 50pix tall with less than 35\% of their body occluded. 

\subsection{16-bit Fixed Point Quantization}
In order to quantize the model, we used Ristretto \cite{ristretto}. Ristretto implements additional layers over caffe in order to train and test a quantized model. We extracted the dynamic range of the parameters and features on each layer, which is basically the maximum absolute value and minimum absolute value. The features were extracted by running the model over 100 images in order to get a dynamic range that most represents the actual one. We fixed all parameters and features bits width to 16-bit, and then we modified the fractional part bits width on each layer based on the corresponding dynamic range. We tested the resulting quantized model and we were able to get similar performance with less than 0.01\% loss in miss-rate. We do not report the speed on the quantized model since Ristretto simulates the fixed-point arithmetic using floating-point arithmetic because there's no native hardware/software support for fixed-point arithmetic on GPUs.

\begin{table}[h!]
	\centering
	\begin{tabular}{|c|c|c|}
	\hline
	 & Miss-rate & Improvement\\
	\hline
	\shortstack{SSD\\(training from scratch)} & 20.32\% & -\\
   	\hline
   	+finetuning & 16.04\% & 4.28\%\\
   	\hline
   	+better hyper-parameters & 11.88\% & 4.16\%\\
   	\hline
   	\end{tabular}
   	\caption{SSD model improvement over the default configuration.}
   	\label{tab:progress}
\end{table}

\begin{table}[h!]
	\centering
\hspace*{-0.25cm}\begin{tabular}{|c|c|c|c|}
	\hline
	Model & \shortstack{Miss-rate\\(Reasonable)} & \shortstack{Miss-rate\\(Overall)} & \shortstack{Speed\\(FPS)}\\
	\hline
	SAF R-CNN & 9.68\% & 62.5\% & 1.7\\
   	\hline
   	MS-CNN & 9.95\% & 60.9\% &15\\
   	\hline
   	CompACT-Deep & 11.75\% & 64.4\% & 2\\
   	\hline
   	RPN+BF & 9.58\% & 64.6\% & 2\\
   	\hline
   	F-DNN & 8.65\% & 50.5\% & 6.25\\
   	\hline
   	SSD-ours & 11.88\% & 54.4\% & 24\\
   	\hline
   	SSD-quantized-ours & 11.89\% & 54.4\% & 24\\
   	\hline
   	\end{tabular}
   	\caption{Comparison between our current model SSD and other state-of-art models on Caltech dataset.}
   	\label{tab:comparison}
\end{table}

\section{Conclusion}
In this paper, we improved Single-Shot Detector on Caltech Pedestrian Detection Benchmark with state of the art performance in terms of detection miss-rate and speed. We also presented a design to accelerate the model using FPGA.

\section{Acknowledgments}
We would like to thank Raghid Morcel for his continued help and support in designing the FPGA accelerator. We would also like to thank Dr. Hazem Hajj, Dr. Mazen Saghir and Dr. Haitham Akkary for guiding through this project, and for their never-ending advice and motivation.

%
%

\nocite{*}
\bibliographystyle{ieeetr}
\bibliography{references}

\newpage


\end{document}